\documentclass[acmsmall]{acmart}
\copyrightyear{2020}
\acmYear{2020}
\setcopyright{acmcopyright}\acmConference[AICCC 2020]{2020 3rd Artificial Intelligence and Cloud Computing Conference}{December 18--20, 2020}{Kyoto, Japan}
\acmBooktitle{2020 3rd Artificial Intelligence and Cloud Computing Conference (AICCC 2020), December 18--20, 2020, Kyoto, Japan}
\acmPrice{15.00}
\acmDOI{10.1145/3442536.3442543}
\acmISBN{978-1-4503-8883-2/20/12}
\AtBeginDocument{%
  \providecommand\BibTeX{{%
    \normalfont B\kern-0.5em{\scshape i\kern-0.25em b}\kern-0.8em\TeX}}}

\begin{document}

\title[Evaluation of Pre-trained CNN models on LC25000 dataset]{Prediction of lung and colon cancer through analysis of histopathological images by utilizing Pre-trained CNN models with visualization of class activation and saliency maps}
\author{Satvik Garg}

\email{satvikgarg27@gmail.com}

\affiliation{%
  \institution{Jaypee University of Information Technology}
  \streetaddress{Waknaghat}
  \city{Solan}
  \state{Himachal Pradesh}
  \country{India}
  \postcode{173234}
}

\author{Somya Garg}
\email{somgarg@deloitte.com}
\affiliation{%
  \institution{Deloitte Consulting LLP}
  \city{New York}
  \country{USA}}


\begin{abstract}
Colon and Lung cancer is one of the most perilous and dangerous ailments that individuals are enduring worldwide and has become a general medical problem. To lessen the risk of death, a legitimate and early finding is particularly required. In any case, it is a truly troublesome task that depends on the experience of histopathologists. If a histologist is under-prepared it may even hazard the life of a patient. As of late, deep learning has picked up energy, and it is being valued in the analysis of Medical Imaging. This paper intends to utilize and alter the current pre-trained CNN-based model to identify lung and colon cancer utilizing histopathological images with better augmentation techniques. In this paper, eight distinctive Pre-trained CNN models, VGG16, NASNetMobile, InceptionV3, InceptionResNetV2, ResNet50, Xception, MobileNet, and DenseNet169 are trained on LC25000 dataset. The model performances are assessed on precision, recall, f1score, accuracy, and auroc score. The results exhibit that all eight models accomplished noteworthy results ranging from 96\% to 100\% accuracy. Subsequently, GradCAM and SmoothGrad are also used to picture the attention images of Pre-trained CNN models classifying malignant and benign images.
\end{abstract}



\maketitle

\section{Introduction}
Cancer is a disease when abnormal cells begin developing wildly, which can start in practically any organ or tissue of the body. Cancer is the driving reason for the second leading cause of death internationally, accounts for an expected 9.6 million passings, or one of every six deaths, in 2018 [1]. Lung cancer (both little cell and non-little cell) represents 2.06 million cases and 1.76 million passings, while colorectal malignancy represents 1.80 million cases and 783K deaths [1].

The two types of cancer in the lungs, which suddenly develop and spread, are small cellular cancer in the lungs (SCLC) and non-small cell cancer (NSCLC). SCLC is a profoundly dangerous tumor from cells showing neuroendocrine qualities and records for 15\% of total lung cancer cases. NSCLC, which represents the staying 85\% of cases, is additionally partitioned into three significant pathology sub-types: adenocarcinoma, squamous cell carcinoma, and enormous cell carcinoma [3]. Colon disease is generally called colorectal cancer, a term that joins colon and rectal cancer, starting in the rectum. Adenocarcinoma account for 96\% of all colorectal cancer cases [4]. In this paper, we mainly focus on analyzing non-small cell lung cancers (NSCLC).

We are automating histopathologists' work, which is to inspect the cells and tissues under a magnifying instrument and distinguishing the irregularity. In recent years, there has been going through a ton of work to catch the whole tissue or cell slide with a scanner and sparing it as a computerized picture. As a result, an enormous number of WSI (whole slide images) are being aggregated. Endeavors have been made to examine WSIs by utilizing machine learning algorithms to diagnose [10]. However, the vast majority of the picture grouping assignments are moved to deep learning since the triumph of the group utilizing deep learning at ImageNet Large Scale Visual Recognition Competition (ILSVRC) 2012 [6].

This research work is summarized as follows: 1) Division of data to perform three binary classification problems. 2) Data augmentation using imgaug [5] toolkit. 3) Feature Extraction using transfer learning through various Pre-trained CNN models [6]. 4) Evaluation of performances on various metrics [7]. 5) Attention visualization using GradCam [8] and SmoothGrad [9]. 6) Conclusion and future work.

\section{Literature Survey}
A good amount of work has done for the classification of histopathological pictures of various cancer types such as breast, lung, colorectal, and skin cancer [10].

Liping Jiao et al. [11] extracted eighteen ordinary features, including grayscale mean, grayscale variance, and 16 texture features extracted by the Gray-Level Cooccurrence Matrix (GLCM) strategy. SVM [12] based classifier was used, which achieved precision, F1-score, recall of 96.67\%, 83.33\%, 89.51\% respectively on 60 colon tissue images partitioned equally into the two classes.

By utilizing textural and architectural-based features, Scott Doyle et al. [13] generated 3400 features from 48 breast tissue images (30 malignant and 18 benign). Moreover, spectral clustering was also performed to decrease the dimensions of the feature set. An SVM [12] classifier is then employed to perform binary classification on cancerous and non-cancerous breast tissue images and distinguish images containing a low and high level of malignancy. Results show that using textural based attributes, 95.8\% accuracy was accomplished in recognizing cancerous tissues from non-malignant growth. By implementing architectural features, 93.3\% accuracy was achieved in recognizing the high level from the low level of cancer.

S. Rathore et al. [14] proposed a feature extraction strategy that mathematically models the constituents of colon tissues' geometric characteristics. Conventional features such as morphological, texture, scale-invariant feature transform (SIFT), and elliptic Fourier descriptors (EFDs) are used to develop a hybrid feature set. SVM [12] is then used as a classifier on 174 colon biopsy images, which results in 98\% accuracy.  

The biggest problem that researchers all over the world are facing is the shortage of data in the clinical field, which results in overfitting [15] of models. To forestall overfitting, F. A. Spanhol et al. [16] applied various data augmentations strategies on BreakHis [18] dataset to increase the number of input images. AlexNet [6] model was then trained from scratch, which results in achieving 90\% accuracy.

H. Chougrad et al. [19], investigating the significance of transfer learning [6], experimentally determine the fine-tuning strategy when training an InceptionV3 [20] model. A larger dataset is created by combining various datasets like the DDSM database [21], the INbreast database [22], BCDR database [23] results in getting an accuracy of about 98.845\%.

Sara Hossein-Zadeh Kassani et al. [27] ensembled well established Pre-trained CNN architectures, namely VGG19 [28], MobileNetV2 [29], and DenseNet201 [30] for aided diagnosis of breast cancer detection on a total of 4 datasets, in particular, BreakHis [18], ICIAR [32], PCam [33], Bioimaging [34] was analyzed. The result shows more than 95\% accuracy is achieved on all datasets aside from the Bioimaging dataset, which attains only 83.1\% accuracy. 

Tetsuya Tsukamoto et al. [35] developed a deep convolutional neural network (DCNN) model for multiclass classification on 298 lung images consisting of three convolutional three pooling layers and two fully connected layers. For preventing overfitting [15], various data augmentation techniques like rotation, zoom, and flip were used. In the results acquired, managing the augmented images, the accuracy of adenocarcinoma, squamous cell carcinoma, and little cell carcinoma is 89.0\%, 60.0\%, and 70.3\%, respectively, which results in a total accuracy of 71.1\%.

From the survey, we examine that a very limited number of studies were done to classify distinctive malignancy subtypes (i.e., to classify different subtypes of lung cancer, colon cancer, breast cancer, etc.). With the accessibility of various Pre-Trained CNN models, the fewer number of models implemented for comparative analysis.
\section{Methodologies}
Figure 1. explains and describes the framework adopted in this research for classification. The framework is divided into 3 phases, namely, Data Preprocessing, Modeling, and Evaluation phase.

\begin{figure*}[htbp]
  \includegraphics[width=0.9\textwidth]{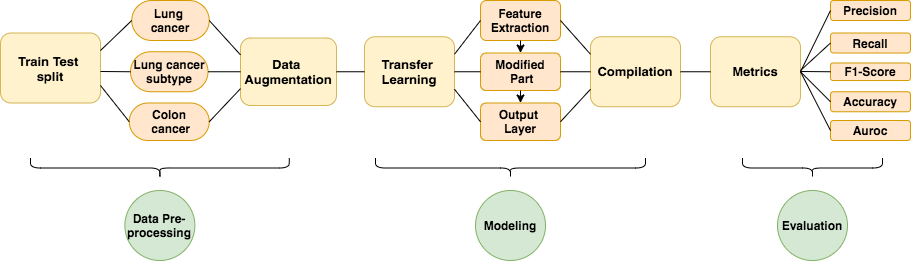}
\caption{Flowchart of the framework adopted}
\end{figure*}

\subsection{Train Test Split}
The dataset used in this research is LC25000 which is known as the lung and colon cancer histopathological image dataset [17]. The dataset was made available to fill the shortage and need of openly accessible datasets in clinical imaging [36]. It contains 15000 lung and 10000 colon pictures partitioned equally into five classes: lung adenocarcinoma, lung squamous cell carcinoma, lung benignant, colon carcinoma, and colon benign. Each class has 5000 images in it, 80:20 split is followed for performing three binary classification tasks:  

a) Binary classification identifying lung cancer in which 5000 images of lung benignant and a total of 5000 images of lung adenocarcinomas and squamous cell carcinomas.

b)	Binary classification between lung cancer subtypes using 10000 pictures of both lung adenocarcinoma and squamous cell carcinoma is used.

c)	Binary classification for finding colon cancer from a set of 5000 colon carcinoma and 5000 non-cancerous colon images. 
 
Table 1 represents the distribution of images for classification used in this research. Column headers of Table 1 are given by; acc/cc = adenocarcinoma/carcinoma, scc = squamous cell carcinoma, ben = benignant.

\begin{table*}[htbp]
\begin{center}
\caption{Overview of the distribution of LC25000 image dataset}
\begin{tabular}{ |p{3.2cm}|p{1cm}|p{1cm}|p{1cm}|p{1cm}|p{1cm}|p{1cm}|  }

 \hline
 \multicolumn{1}{|c|}{Classification task} & \multicolumn{3}{|c|}{Training set 80\%} & \multicolumn{3}{|c|}{Test set 20\%} \\
 \cline{2-7} 
 & \,acc\textbackslash cc & \,\,\,\,scc & \,\,\,\,ben & \,acc\textbackslash cc & \,\,\,\,scc & \,\,\,\,ben \\
\hline
Lung cancer&\,\,\,2000    &\,\,\,2000& \,\,\,4000& \,\,\,\,500&\,\,\,\,500&\,\,\,1000\\
 \hline
Lung cancer subtypes& \,\,\,4000&\,\,\,4000& \quad\,--& \,\,\,1000&\,\,\,1000&\quad\,--\\
\hline
Colon cancer&\,\,\,4000    &\quad\,--& \,\,\,4000& \,\,\,1000&\quad\,--&\,\,\,1000\\
  \hline

\end{tabular}
\end{center}
\end{table*}%


\subsection{Data Augmentation}
The images acquired from LC25000 [17] are of size 768*768. All images are cropped to square sizes of 224 x 224 pixels from the original 768 x 768 pixels. The images, in the LC25000 image dataset, were already augmented by the following augmentations: left and right rotations (up to 25 degrees, 1.0 probability) and by horizontal and vertical flips (0.5 probability) [17]. However, a complex augmentation pipeline from the imgaug library [5] is used in this research to boost the performance of Pre-trained CNN models. An original as well as a collection of 64 augmented images, using the imgaug library [5], of a colon and lung adenocarcinoma from the LC25000 image dataset is given in Fig. 2. respectively.

\begin{figure*}[htbp]
  \includegraphics[width=0.1\textwidth]{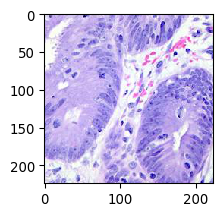}
  \includegraphics[width=0.37\textwidth]{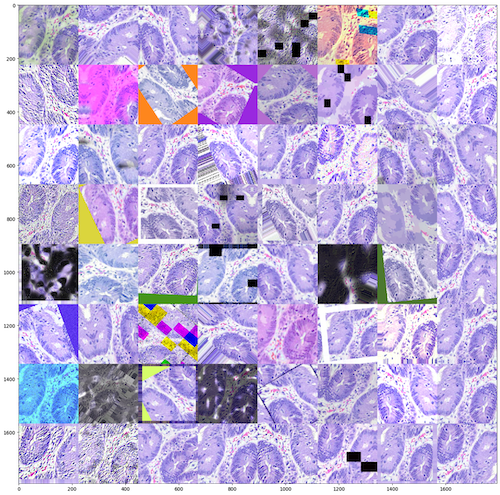}
  \includegraphics[width=0.1\textwidth]{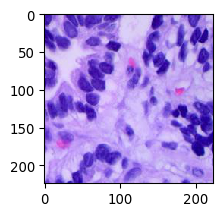}
  \includegraphics[width=0.37\textwidth]{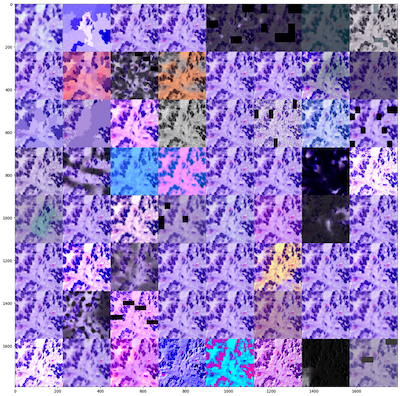}
\caption{Original and 64 augmented images using imgaug library of both colon and lung carcinoma respectively.}
\end{figure*}

\subsection{Transfer Learning}
With an expanding number of clients on the web, there is a vast amount of information accessible for analysis. In the coming years, it would be tough to deal with such enormous amounts of data [31]. Unfortunately, the clinical field datasets are very scant because of protection and privacy concerns [36]. On the other hand, histological images are very unpredictable. It would be difficult and troublesome to prepare a model without any preparation from scratch with arbitrarily initialized features with low data measures.

To handle these issues, Pre-trained CNN models [6] are used in this research. The idea is to transfer the weights of a pre-existing model to other problems. This is also known as transfer learning [37]. It aids in lessening the computational force and accelerates the system's performance, which brings about showing signs of faster and improved results. In this research, we propose a structure that would apply to all three binary classification tasks for training our model. The structure contains three sections:

a)	Feature Extraction Part - Extracting the most important features by fine-tuning the models through training the eight notable Pre-trained CNN architectures. These features after this point would go to the modified part for further classification.

b)	Modified Part - Added a concatenate layer which is a mix of three layers that consolidates Max-pooling2D, Average-pooling2D, and a flatten layer to get a vector of features. Besides, a dropout layer with a dropout rate of 0.5 has also been incorporated.

c)	Output Section - Added a vector of features to an output layer with a sigmoid activation function which is expressed as follows:    \begin{equation}
f(x) = \frac{1}{1+e^{-x}}
\end{equation}

\subsection{Compilation}
The compilation of the model requires an optimization algorithm. There are various optimization algorithms such as adaptive learning rate (Adam), stochastic gradient descent (Sgd), Adadelta, RMSProp, AdaMax, and many more. Adam is chosen as it is the standard most and works commendably well for any kind of deep learning model [41]. The hyperparameters values used for training the models are:  Learning rate = 0.0001, epochs = 10, Batch size = 64.

\subsection{Metrics}
The models are dissected on precision, recall, f1-score, accuracy, and auroc score which assists with picking the best model. Let the letter be: 

$(T_p)$ = Number of times model predicts the positive class correctly.

 $(T_n)$ = Number of times model predicts the negative class correctly.
 
$(F_p)$ = Number of times model predicts the positive class incorrectly.

$(F_n)$ = Number of times model predicts the negative class incorrectly.

Precision, recall, accuracy, and f1score be expressed as follows:
\begin{equation}
Precision = \frac{T_{p}}{T_{p}+F_{p}}
\end{equation}
\begin{equation}
Recall = \frac{T_{p}}{T_{p}+F_{n}}
\end{equation}
\begin{equation}
Accuracy = \frac{T_{p}+T_{n}}{T_{p}+T_{n} + F_{p}+F_{n}}
\end{equation}
\begin{equation}
F1-Score = 2.\frac{Precision.Recall}{Precision+Recall}
\end{equation}

Auroc is an area under a region operating curve which shows the relation between true positive rate and false positive rate across various thresholds.

\section{Results}
In this experiment, eight different Pre-trained CNN models are used to perform three binary classification tasks, i.e., the classification between lung cancer subtypes, identifying malignant and benign colon as well as lung histological images. Test data is utilized for test predictions that represent 20 percent of the total data. Precision, Recall, F1-score, Accuracy, and Auroc score are assessed to evaluate model performance. Training loss, validation loss, training accuracy, and validation accuracy with respect to epochs are also plotted to get a visual analysis of model performance while training the test and validation data. The validation data used is 20\% of the rest of the training data (i.e., after 80:20 train test split).

\subsection{Lung cancer}
Evaluation metrics of all eight models are given in Table 2. Predictions shows that all eight Pre-trained CNN designs attained remarkable results, 100\% in precision, accuracy, recall, f1score, and auroc score. Notwithstanding, there is an exclusion that the NASNetMobile model has accomplished 99\% auroc score. 

\begin{table*}[htbp]
\caption{Evaluation metrics of eight Pre-trained CNN models to identify lung cancer}
\begin{tabular}{ |c|c|c|c|c|c| } 
\hline
\,Model\, & \,Precision\, & \,Recall\, & \,F1-score\, & \,Accuracy & \,Auroc \,  \\
\hline
 VGG16 & 1.0 & 1.0 & 1.0 & 1.0 & 1.0 \\ 
 ResNet50  & 1.0 & 1.0 & 1.0 & 1.0 & 1.0\\ 
 InceptionV3 & 1.0 & 1.0 & 1.0 & 1.0 & 1.0\\ 
 \,InceptionResNetV2\,  & 1.0 & 1.0 & 1.0 & 1.0 & 1.0\\ 
 MobileNet  & 1.0 & 1.0 & 1.0 & 1.0 & 1.0 \\
 Xception  & 1.0 & 1.0 & 1.0 & 1.0 & 1.0\\ 
 NASNetMobile  & 1.0 & 1.0 & 1.0 & 1.0 & 0.99\\
 DenseNet169  & 1.0 & 1.0 & 1.0 & 1.0 & 1.0\\
\hline
\end{tabular}
\end{table*}

Analyzing figure 3 shows that besides ResNet50 all models are indicating comparable plots bringing about early convergence. ResNet50 in the last 2-3 epochs showed a sudden improvement in limiting loss and maximizing accuracy.

\begin{figure*}[htbp]
  \includegraphics[width=3.3cm, height=2.9cm]{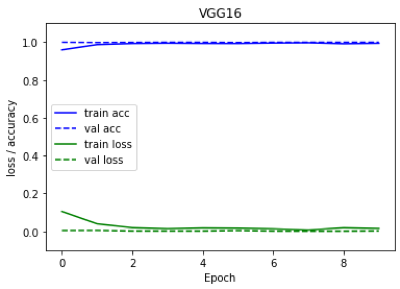}
  \includegraphics[width=3.3cm, height=2.9cm]{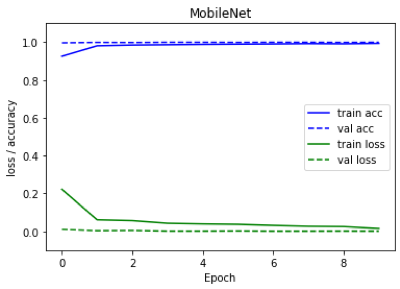}
  \includegraphics[width=3.3cm, height=2.9cm]{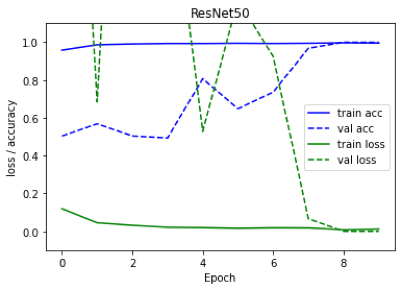}
  \includegraphics[width=3.3cm, height=2.9cm]{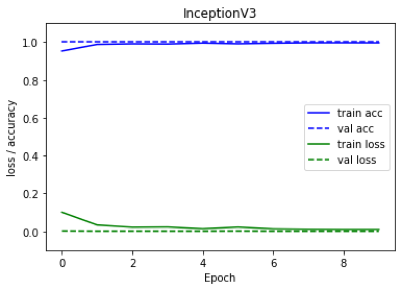}
  \includegraphics[width=3.3cm, height=2.9cm]{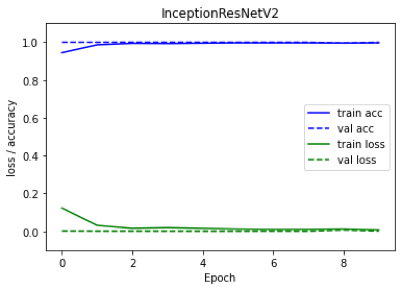}
  \includegraphics[width=3.3cm, height=2.9cm]{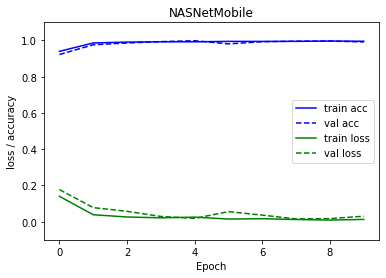}
  \includegraphics[width=3.3cm, height=2.9cm]{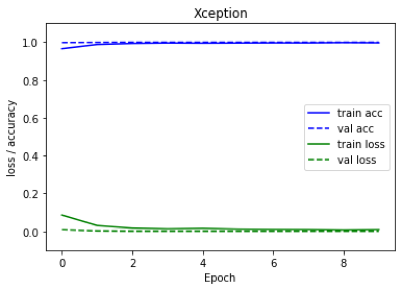}
  \includegraphics[width=3.3cm, height=2.9cm]{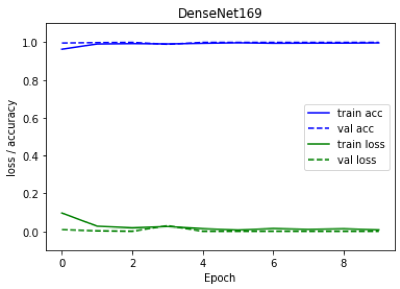}
\caption{Change in loss and accuracy of training and validation data with respect to the number of epochs of eight Pre-trained CNN models classifying lung malignant and benign images.}
\end{figure*}

\subsection{Lung cancer subtypes}
This is a classification between lung cancer subtypes (lung carcinoma and lung squamous cell carcinoma). Table 3 summarize that InceptionV3, InceptionResnetV2, DenseNet169, MobileNet, and Xception accomplished 100\% outcomes in all evaluation metrics aside from the auroc score in which MobileNet and DenseNet169 accomplished 99.9\% score. The remaining models VGG16, Resnet50, and NASNetMobile accomplished marginally lesser results ranging from 96-98

\begin{table*}[htbp]
\caption{Evaluation of eight Pre-trained CNN models identifying type of lung cancer}
\begin{tabular}{ |c|c|c|c|c|c| } 
\hline
\,Model\, & \,Precision\, & \,Recall\, & \,F1-score\, & \,Accuracy & \,Auroc \,  \\
\hline
 VGG16  & 0.975 & 0.975 & 0.98 & 0.98 & 0.999 \\ 
 ResNet50  & 0.965 & 0.965 & 0.96 & 0.96 & 0.999\\ 
 InceptionV3  & 1.0 & 1.0 & 1.0 & 1.0 & 1.0 \\ 
 ,InceptionResNetV2\,  & 1.0 & 1.0 & 1.0 & 1.0 & 1.0\\ 
 MobileNet  & 1.0 & 1.0 & 1.0 & 1.0 & 0.999 \\
 Xception  & 1.0 & 1.0 & 1.0 & 1.0 & 1.0\\ 
 NASNetMobile  & 0.965 & 0.965 & 0.97 & 0.97 & 0.997 \\
 DenseNet169  & 1.0 & 1.0 & 1.0 & 1.0 & 0.999\\
\hline
\end{tabular}
\end{table*}

Figure 4 depicts that NASNetMobile and ResNet50 are not in an arrangement with other models i.e. the dotted green validation loss curve is slightly on upward side. In addition, ResNet50 again showed uncertainty, followed a similar pattern as it was classifying lung images.
\begin{figure*}[htbp]
  \includegraphics[width=3.3cm, height=3cm]{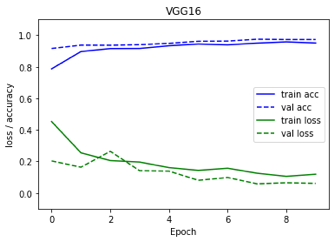}
  \includegraphics[width=3.3cm, height=3cm]{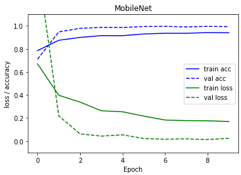}
  \includegraphics[width=3.3cm, height=3cm]{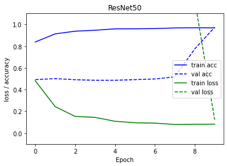}
  \includegraphics[width=3.3cm, height=3cm]{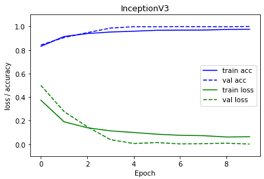}
  \includegraphics[width=3.3cm, height=3cm]{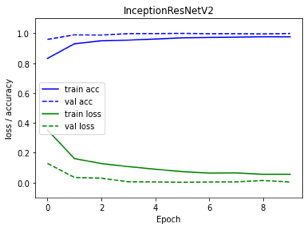}
  \includegraphics[width=3.3cm, height=3cm]{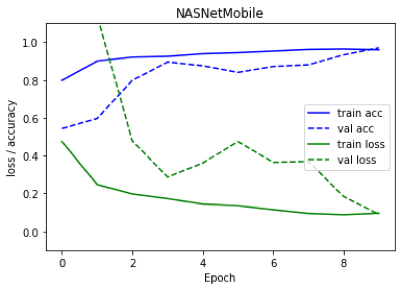}
  \includegraphics[width=3.3cm, height=3cm]{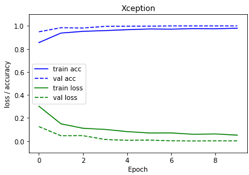}
  \includegraphics[width=3.3cm, height=3cm]{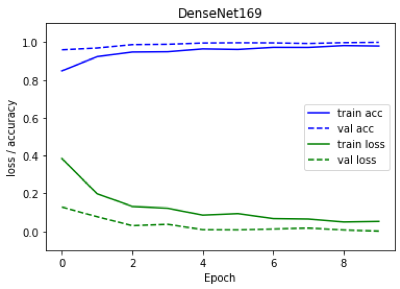}
\caption{Change in loss and accuracy of training and validation data with respect to the number of epochs of eight Pre-trained CNN models classifying the type of lung cancer.}
\end{figure*}

\subsection{Colon Cancer}
Table 4 sums up that all models performed exceptionally well, attaining 100\% accuracy in all evaluation metrics. However, NASNetMobile scored around 98\% accuracy. For the most part, these are above par results.

\begin{table*}[htbp]
\caption{Evaluation metrics of Pre-Trained CNN model classifying colon cancer images}
\begin{tabular}{ |c|c|c|c|c|c| } 
\hline
\,Model\, & \,Precision\, & \,Recall\, & \,F1-score\, & \,Accuracy & \,Auroc \,  \\
\hline
 VGG16  & 1.0 & 1.0 & 1.0 & 1.0 & 1.0 \\ 
 ResNet50  & 1.0 & 1.0 & 1.0 & 1.0 & 1.0\\ 
 InceptionV3  & 1.0 & 1.0 & 1.0 & 1.0 & 1.0 \\ 
 \,InceptionResNetV2\, & 1.0 & 1.0 & 1.0 & 1.0 & 1.0\\ 
 MobileNet  & 1.0 & 1.0 & 1.0 & 1.0 & 1.0\\
 Xception  & 1.0 & 1.0 & 1.0 & 1.0 & 1.0\\ 
 NASNetMobile  & 0.98 & 0.98 & 0.98 & 0.98 & 1.0\\
 DenseNet169  & 1.0 & 1.0 & 1.0 & 1.0 & 1.0 \\
\hline
\end{tabular}

\end{table*}

Figure 5 manifests that ResNet50 likewise displayed a similar type of plot that appeared in lung classification model. The 2\% decrease in evaluation metrics of NASNetMobile can easily be depicted by the zigzag nature of validation loss and accuracy curve of NASNetMobile architecture shown in figure 5. 
\begin{figure*}[htbp]
  \includegraphics[width=3.3cm, height=3cm]{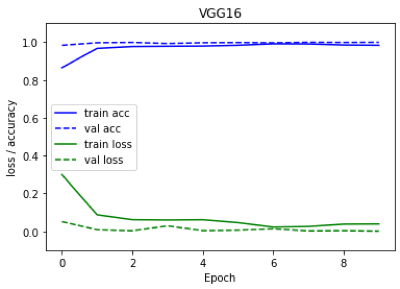}
  \includegraphics[width=3.3cm, height=3cm]{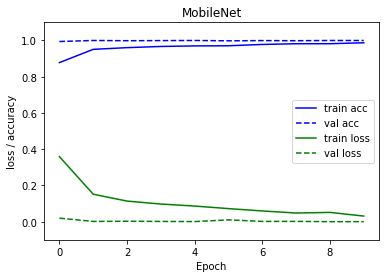}
  \includegraphics[width=3.3cm, height=3cm]{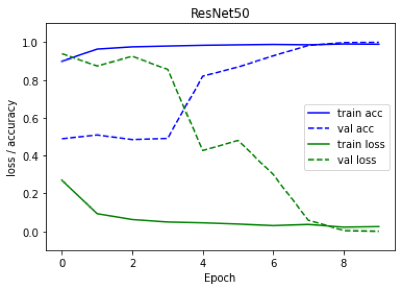}
  \includegraphics[width=3.3cm, height=3cm]{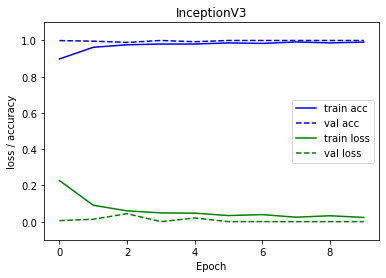}
  \includegraphics[width=3.3cm, height=3cm]{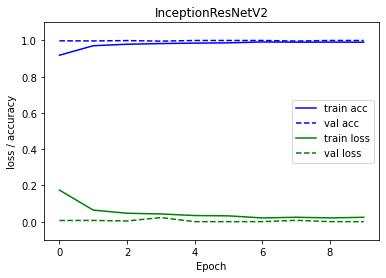}
  \includegraphics[width=3.3cm, height=3cm]{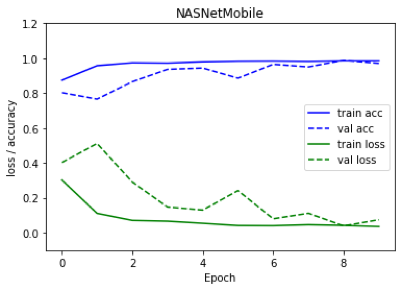}
  \includegraphics[width=3.3cm, height=3cm]{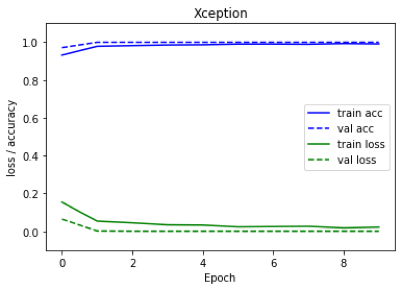}
  \includegraphics[width=3.3cm, height=3cm]{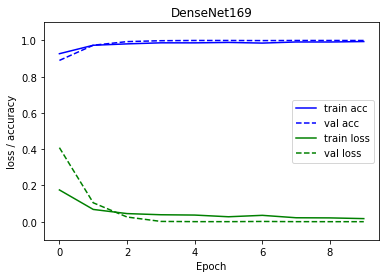}
\caption{Change in loss and accuracy of both training and validation data with respect to the number of epochs of eight Pre-trained CNN models classifying colon malignant tissues from benign ones.}
\end{figure*}

\subsection{Discussion}
The results acquired from all the binary classification tasks are acceptable, yet that doesn't imply that these models are prepared for applications in real life. The models still need improvements. The dataset used in this research comes from a single source only, i.e., LC25000. We need to combine distinctive datasets to make a summed-up model. This paper intends to show just the methodology used to perform binary classification in histopathology images. However, we achieved good accuracy in classifying different malignancy subtypes, which shows that these Pre-trained CNN models own the ability to do multiclass classification on various cancer types using Histopathology images with quality results.  

\section{Attention Visualization}
Visualizing activation maps helps in featuring those areas of the image which are important, where activation is high. It gives users certainty that their work is going the correct way. In this paper, GradCam and SmoothGrad are used to visualize class activation and saliency maps respectively. Saliency Maps such as SmoothGrad and vanilla saliency [25] uses the gradients of the output layer with respect to input image which tells us how the output value changes with a change in input value whereas class activation maps [24] such as GradCAM, ScoreCAM, and many more, didn't use gradients instead they use penultimate layer (pre-dense) to get spatial information which gets lost in dense layers. tf-keras-vis [42] visualization toolkit was used for visualization of these maps.

Figure 6. shows the gradient-based class activation maps and saliency maps using GradCAM and SmoothGrad for visualizing attention by utilizing NASNetMobile model for the lung and VGG16 model for colon categorizing cancerous and noncancerous images. 

\subsection{GradCAM}
GradCAM, unlike CAM, uses the gradients of target concept which then flows into the final convolutional layer of CNN, and then result in obtaining localization map $L^c_{Grad-CAM} \; \epsilon \; R^{u\times v}$ of width $u$ and height $v$ for any class $c$, requires to obtain neuron importance weight which can be calculated by computing the gradient of the score for class before softmax ($y^c$) to the feature map activation ($A^k$) of a convolutional layer, i.e. $\frac{\partial y^c} {\partial A^k}$ [8].
\begin{align}
\alpha^c_k &= \frac{1}{Z}\,\Sigma_i\,\Sigma_j\,\frac{\partial y^c}{\partial A^k_{ij}}\\
L_{Grad-CAM}^c& = ReLU\left(\Sigma_{k}\,\, \alpha^c_kA^k\right)
\end{align}

\begin{figure*}[htbp]
  \includegraphics[width=0.45\textwidth]{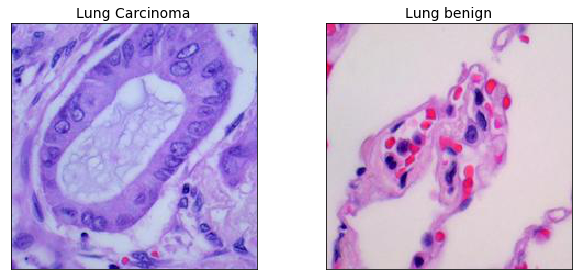}
  \quad\includegraphics[width=0.45\textwidth]{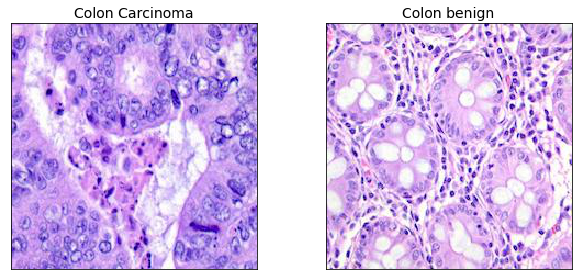}
  \includegraphics[width=0.45\textwidth]{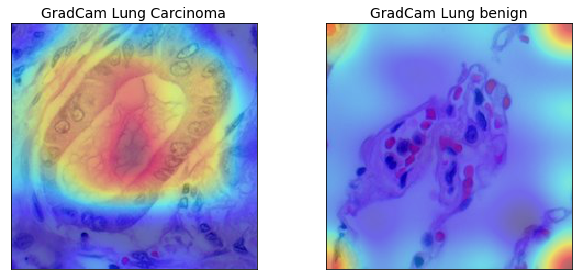}
  \quad\includegraphics[width=0.45\textwidth]{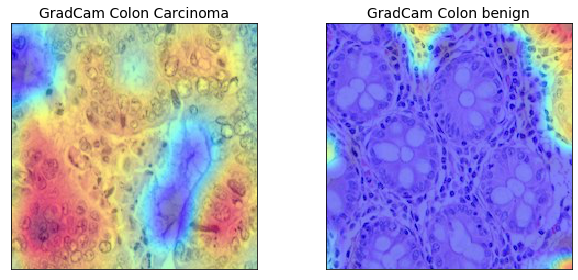}
  \includegraphics[width=0.45\textwidth]{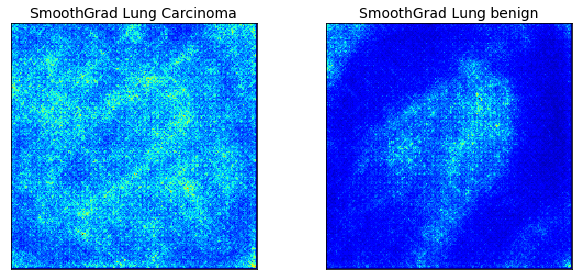}
  \quad\includegraphics[width=0.45\textwidth]{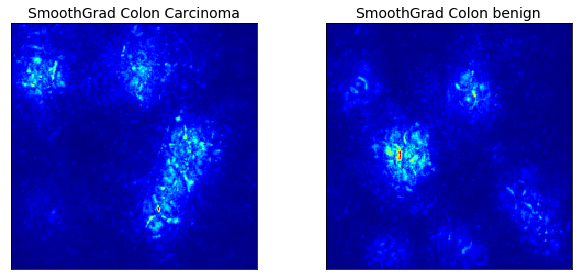}
\caption{Attention visualization of images employing GradCAM for activation maps and SmoothGrad for saliency maps of both lung and colon image classifying carcinoma(cancerous) with benign(non-cancerous) images.}
\end{figure*}

\subsection{SmoothGrad}
Saliency Maps can be visualized by vanilla saliency or SmoothGrad. SmoothGrad was adopted in this work as the vanilla saliency map is too noisy. The idea of the SmoothGrad is to enhance the saliency maps by adding noise to the input image [9].

For an input image $x$, computing a class activation function $S_{c}$ for each class $c\,\epsilon\, C$, the final classification $class(x)$ is:
\begin{equation}
    class(x) = argmax_{c\,\epsilon\, C}\,\, S_{c}(x)
\end{equation}

Sensitivity map $M_c(x)$ can be calculated by differentiating $S_c$ with respect to input $x$.
\begin{equation}
    M_c(x) = \partial S_c(x) / \partial x
\end{equation}

It has been shown that the gradient of $S_c$ is producing fluctuations rapidly [9]. To improve sensitivity maps, a neighborhood average of gradient values has been used on smoothing of $\partial{S_c}$ with a Gaussian kernel. The smoothed gradient $\hat{M_c(x)}$ for an input image $x$, is represented by:
\begin{equation}
    \hat{M_c(x)} = 1/n\,\Sigma_{i}^n M_c(x + N(0,\sigma^2))
\end{equation}

where $n$ is the number of samples, $N(0,\sigma^2)$ is Gaussian noise with standard deviation $\sigma$ and $M_c$ represents unsmoothed gradient [9].

\section{Conclusion}
In this research, we discussed the need for automation in clinical imaging, which can radically diminish the heap from specialists. Then we present related works that go from using basic models to advance models like deep learning. At that point, we played out with some information enlargement strategies on our histology pictures. We applied eight popular Pre-trained CNN models, VGG16, ResNet50, MobileNet, InceptionResNetV2, NASNetMobile, Xception, InceptionV3, DenseNet169, which accomplished remarkable outcomes ranging from 97.5\% to 100\% accuracy specifically InceptionResNetV2, InceptionV3, MobileNet got 100\% precision, accuracy, auroc score, recall, f1-score in all classification tasks. All remaining models have also demonstrated exceptional results ranging from 97.5\% to 99.5\% in all evaluation parameters. These high accuracy results from applying a complex image augmentation pipeline (imgaug) combined with complex Pre-trained CNN architectures. Training and validation loss and accuracy of various Pre-trained CNN models have also been envisioned, which shows that ResNet50 is demonstrating unexpected conduct compared to different models. Besides, class activation and saliency maps have been pictured by utilizing GradCAM and SmoothGrad to imagine the model's consideration layers. In the future, we will apply all the discussed methods in classifying other cancer types by merging different datasets and do multi-class classification between different malignancy subtypes.    

\bibliographystyle{ACM-Reference-Format}

\end{document}